\documentclass{article}
\usepackage{ijcai25}

\usepackage{times}
\usepackage{soul}
\usepackage{url}
\usepackage[hidelinks]{hyperref}
\usepackage[utf8]{inputenc}
\usepackage[small]{caption}
\usepackage{graphicx}
\usepackage{amsmath}
\usepackage{amsthm}
\usepackage{booktabs}
\usepackage{algorithm}
\usepackage{algorithmic}
\usepackage[switch]{lineno}

\usepackage{multirow}
\usepackage{diagbox}

\usepackage{xcolor}

\usepackage{amsfonts}

\title{
Towards Machine Learning-based Model Predictive Control for HVAC Control in Multi-Context Buildings at Scale via Ensemble Learning
}

\author{
    Yang Deng$^1$, Yaohui Liu$^1$, Rui Liang$^1$, Dafang Zhao$^2$, Donghua Xie$^1$,  \\ Ittetsu Taniguchi$^2$ and Dan Wang$^1$
    \affiliations
    $^1$Department of Computing, The Hong Kong Polytechnic University, Hong Kong SAR \\
   $^2$Graduate School of Information Science and Technology , Osaka University, Osaka, Japan
    \emails
    {marco.deng, dan.wang}@polyu.edu.hk, zhao.dafang@ist.osaka-u.ac.jp
}

\begin{document}

\maketitle

\begin{abstract}


%
%






The building thermodynamics model, which predicts real-time indoor temperature changes under potential HVAC (Heating, Ventilation, and Air Conditioning) control operations, is crucial for optimizing HVAC control in buildings. While pioneering studies have attempted to develop such models for various building environments, these models often require extensive data collection periods and rely heavily on expert knowledge, making the modeling process inefficient and limiting the reusability of the models.
This paper explores a \textit{model ensemble} perspective that utilizes existing developed models as base models to serve a target building environment, thereby providing accurate predictions while reducing the associated efforts. Given that building data streams are non-stationary and the number of base models may increase, we propose a Hierarchical Reinforcement Learning (HRL) approach to dynamically select and weight the base models.
Our approach employs a two-tiered decision-making process: the high-level focuses on model selection, while the low-level determines the weights of the selected models.
We thoroughly evaluate the proposed approach through offline experiments and an on-site case study, and the experimental results demonstrate the effectiveness of our method.

\end{abstract}

\section{Introduction}


According to government statistics, residential energy consumption accounts for about 40\% of society’s total energy consumption. The International Energy Agency (IEA) predicts that by 2050, this amount will increase by 50\%.
Recent IEA reports~\cite{IEAreport} indicate that about half of building energy use is attributed to HVAC (heating, ventilation, and air conditioning) systems. This significant energy consumption arises from the substantial demands placed on heating and cooling systems to transfer heat between indoor and outdoor environments, a process governed by building thermodynamics. By implementing optimal HVAC control without compromising occupant thermal comfort, energy consumption in buildings can be significantly reduced.

Recently, the control optimization of HVAC systems has evolved from traditional methods (e.g., bang-bang control, feedback-based control) to Model Predictive Control (MPC), a constrained optimal control strategy that determines the optimal control inputs by minimizing a predefined objective function~\cite{yao2021state}. For instance, an electrical price-aware MPC scheme for floor heating systems was developed for a residential apartment in Denmark~\cite{hu2019price}.
As the name suggests, the MPC scheme requires appropriate modeling of systems and components to effectively control the dynamic processes within the building and HVAC system. The model serves as the cornerstone of the MPC system, enabling the prediction of future thermodynamics—typically reflected by indoor temperature\footnote{We emphasize that predicting thermodynamics is analogous to predicting indoor temperature, a practice commonly found in the literature.}—of the building environment.
A schematic diagram of MPC in building HVAC systems is presented in figure. Various modeling schemes have been developed for the associated environments, including both physics-based and data-driven approaches~\cite{yao2021state}. With a more accurate thermodynamic model, the effectiveness of control actions derived from optimization can be significantly enhanced.

However, while models have been developed for different building environments and control systems, several challenges persist. Among these, the primary ones are: 
(1) The modeling process highly involves expertise and experimental studies. 
The input and output variables of the model should be quantified as a mathematical relationship and thus support the optimizer solver tool, thus the majority of models are white-box or gray-box. Selecting suitable variables usually depends on the building's properties~\cite{drgovna2020all}.
(2) The model needs sufficient data collection. All the modeling methods need data to construct and calibrate the model, especially the data-driven models.

To solve these challenges, we leverage the model ensemble paradigm~\cite{yang2023survey} from ML community, which has proven effective in enhancing prediction performance across various tasks, including autonomous driving and climate forecasting~\cite{kim2022learning,fu2022reinforcement}. 
Ensemble modeling combines multiple base models to reduce generalization errors, with recent applications in building HVAC systems, including chiller fault detection~\cite{yao2022fault}.
In this context, modeling a specific building environment can be achieved through an ensemble of models developed for other environments. However, selecting and weighting the base models in the ensemble remains an open question, as a suboptimal ensemble can prevent the models from reaching their full potential. This is particularly important for two reasons:
i) The model requires calibration to accommodate changes in context resulting from variations in indoor behavior, as well as mode switches, such as transitioning from cooling to heating.
ii) The number of base models may increase significantly through crowdsourcing, particularly as the building automation community embraces the ensemble paradigm.

In this work, we investigate model ensemble for thermodynamic modeling of buildings as a reinforcement learning (RL) problem. RL has proven effective for decision-making in model selection and data selection~\cite{fu2022reinforcement,zhang2022multi,qiu2021reinforced}. 
It is particularly appealing for our task because it can learn the ensemble policy solely from logged data, without requiring knowledge of the underlying physics in this domain.
We propose a Hierarchical Reinforcement Learning-based approach named \textit{ReeM}. \textit{ReeM} is a two-level decision-making process: in the high-level decision-making phase, an agent determines whether a model should be included in the ensemble; in the low-level decision-making phase, the agent assigns weights to each selected base model. Separate rewards are designed to encourage the two agents to focus on their respective tasks. Furthermore, to enable effective actions, a data representation module is implemented to extract relevant information from the observed system states.
We analyze the prediction performance of \textit{ReeM} across multiple rooms in a real building and conduct an on-site case study to demonstrate its feasibility in a live HVAC control task. The main contributions of this work are summarized as follows:

\begin{itemize}

\item We introduce an ensemble perspective for the first time to enhance the efficiency of building thermodynamic model development. 

\item We propose a novel ensemble approach based on a hierarchical reinforcement learning framework specifically designed for the building model ensemble task.

\item We conduct offline experiments that demonstrate our approach significantly improves performance compared to existing schemes, resulting in an accuracy improvement of 44.54\% over the customized model and 51.65\% over the ensemble baselines. 
Additionally, our on-site case study shows that our approach achieves a reduction in energy consumption compared to the baseline.


\item We make our code and part of datasets available to the AI community to catalyze research in this problem: https://anonymous.4open.science/r/ReeM-2025.

\end{itemize}

\section{Background and Related Work}


\textbf{HVAC control optimization.}
HVAC control optimization refers to the process of utilizing data and control technologies to enhance the performance of building HVAC systems. The primary objectives are to improve energy efficiency, reduce costs, and enhance occupant comfort. Various control strategies have been implemented in HVAC systems, including fuzzy logic control~\cite{berouine2019fuzzy}, genetic algorithms~\cite{huang1997using}, and model predictive control (MPC)~\cite{zhao2023data}.
Recently, MPC is commonly deployed in modern HVAC control and stands out among these strategies due to its solid theoretical foundation and dynamic optimization capabilities.
Unlike other strategies that primarily rely on rules or heuristic approaches, MPC anticipates future system behavior prediction by modeling the thermodynamics of the building environment over a specified horizon, effectively generating a series of what-if predictions. 
Based on this model, MPC determines the optimal control actions—such as setpoints or operational statuses of HVAC equipment—that minimize a predefined cost function while satisfying system constraints, such as indoor temperature limits.
This work focuses on the optimization of HVAC control based on MPC\footnote{For the sake of simplicity, we will refer to it as ``HVAC control'' in the remainder of this paper.}, and the standard form of the optimization is as follows: where $J$ represents the objective function, $f()$ denotes the prediction model. The three parameters (i.e., state, control input, disturbance) for modeling will be introduced later.
\begin{equation}
\begin{aligned}
\min_{u, x} \quad & \sum_{t=0}^{N-1} J(x_{t+1}, u_t) \\
\text{s.t.} \quad & x_{t+1} = f(x_{t-}, u_{t-}, d_{t-}) 
, \quad (x, u) \in (\mathcal{X}, \mathcal{U}) 
\end{aligned}
\end{equation}



\textbf{Thermodynamics modeling of building.}
The thermodynamics model is the cornerstone of HVAC control optimization and must satisfy the requirements~\cite{yao2021state}: 
(i) it should accurately describe system dynamics for a given set of variables, and (ii) it must be computationally tractable to allow for real-time optimization. The modeling techniques in the literature can be categorized into physical and data-driven models. In HVAC control scenarios, classical physical models include the Resistance-Capacitance (RC) Model
~\cite{afram2014review} and the Thermal Equivalent Circuit Model (TECM)~\cite{gan2020development}. Both models describe the transfer and storage of heat using simplified circuit theory and are commonly employed in energy consumption simulation and control system design.
On the other hand, data-driven methods~\cite{leprince2021fifty,ding2022study} encompass a wide range of models that include simplified physical relationships while also requiring parameter estimation based on measured data. 
For example, a symbolic regression (SR) model ~\cite{leprince2021fifty} has been developed to model building thermodynamics in both cooling and heating modes. It is important to note that within the data-driven category, black-box schemes, such as deep neural networks~\cite{qi2024spatio}, have been employed to predict indoor temperature changes. However, these models are challenging to apply directly to HVAC control optimization due to their lack of interpretability and the absence of explicit dynamic relationships between inputs and outputs.
Therefore, the practical model for HVAC control is typically expressed as a mathematical relationship, as shown in Eq.~\ref{definition: M} from the literature~\cite{yao2021state,drgovna2018approximate}, where $x$, $u$ and $d$ are states, control inputs and disturbances, respectively. And '-' denotes the moving average terms.
\begin{equation}
    \label{definition: M}
   f(\cdot,\cdot, \cdot) = \underbrace{Ax_{t-} + B_1u_{(t-1)-}+Ed_{t-}}_{\text{observed part}} + \underbrace{B_2u_{t}}_{\text{control variable}}
\end{equation}



Note that, the implementation of Eq.~\ref{definition: M} can be complex, since $x$, $u$, and $d$ could involve multiple variables~\cite{yufei2024data,cui2019hybrid}.
While these approaches demonstrate strong performance, existing thermodynamic modeling techniques rely on extensive data collection from the target building environment and often require expert analysis, which renders the modeling process inefficient.

\textbf{Ensemble of models.}
Learning an ensemble of base models has become a popular approach in machine learning to enhance overall task performance, reduce prediction variance, and prevent overfitting~\cite{ovadia2019can,yang2023survey}. In data streaming scenarios, determining the weights for base models can be particularly challenging due to the dynamic and non-stationary nature of the data distribution. Early works have employed hand-crafted rules or heuristics~\cite{cerqueira2017dynamic} to address this issue.
Another line of research formulates the ensemble problem as a reinforcement learning (RL) problem~\cite{fu2022reinforcement}. For instance, a previous study~\cite{lin2023deep} utilizes the Actor-Critic framework to create a dynamic ensemble model for stock prediction.

\section{Problem Statement}

For a target building environment (which can refer to either a building or room level), HVAC control optimization heavily relies on the building thermodynamics model to achieve what-if indoor temperature predictions. 
The three major classes of elements are as follows:
1) \textit{Building states} $\mathbf{x_{t}} \in \mathbb{R}^{L}$: indoor temperatures representing the status of the system;
2) \textit{Control inputs} $\mathbf{u_{t-1}} \in \mathbb{R}^{L-1}$: powers of HVAC systems which serve as control signals to the system;
3) \textit{Disturbances} $\mathbf{d_{t}} \in \mathbb{R}^{L \times c}$: features including ambient temperature and occupancy behavioral patterns describing external influences.
Here, $L$ is the look-back window length. 

To model a given environment $e$, existing strategies typically develop a specific function $f_e$ for that environment, as illustrated in Fig~\ref{fig: story}(a). 
In this paper, we revisit this problem from a model ensemble perspective, as shown in Fig~\ref{fig: story}(b). Specifically, given the features $\mathbf{x_{t}}$, $\mathbf{u_{t-1}}$, and $\mathbf{d_{t}}$ collected at each time step $t$, along with a set of $N$ customized models $\{f_{i}\}_{i=1}^{N}$ developed for diverse building environments, our goal is to automatically construct a model ensemble $f^{ens}_t$ to achieve the best prediction performance by
determining the optimal weight vector $\mathbf{w}=[w_{1},...,w_{N}]$ for the base models. The ensemble model is defined as: $f_{t}^{ens} = \sum_{i=1}^{N}w_{i}f_{i}$ and the predicted state at time $t+1$ is given by $\hat{x}_{t+1} = f_{t}^{ens}(\cdot, \cdot, \cdot)$ when input a control $u_t$ at time $t$.
The problem can be formulated as:
$\mathbf{w}^{*} = \arg\min_{\mathbf{w}} \mathcal{L}(f_{t}^{ens})$, where $\mathcal{L}$ denotes the prediction error.

\section{Method}

In this section, we introduce the architecture of our hierarchical RL-based  model ensemble approach \textit{ReeM},
including MDP formulation, state representation, high-level agent, and low-level agent.


\subsection{MDP Formulation}



Considering the extensive search space resulting from the large number of base models and the non-stationary system states, we describe the task of ensemble modeling in this context as a hierarchical Markov Decision Process (MDP). Specifically, we decompose the overall MDP task $M$ into two classes of subtasks: $M_{h}$ and $M_{l}$ for model selection and model weight search, respectively. 
Both subtasks are defined as $\left \langle S, A, P, r, \gamma \right \rangle  $, where $S$ is the state space and $A$ is the action space. $P$ represents the transition function, $r$ is the reward, and $\gamma$ is the discount factor.
In our scenario, the transition $P(s_{t+1} |s_t, a_t) = P(s_{t+1} |s_t)$ is irrelevant to the output model ensemble, as $a_t$ does not influence the next state $s_{t+1}$. Generally, since the model in the MPC context conduct one-step prediction, we can set $\gamma=0$.

\begin{itemize}
    \item \textit{High-level state} $s^{h}_{t}$ at time $t$ contains historical thermal status, control records, and external disturbances of the target building, as well as the prediction performance of base models from the model library $\{f_{i}\}_{i=1}^{N}$.

    \item \textit{High-level action} $a^{h}_{t}$ is a binary vector indicating the base models selected at time $t$.

    \item \textit{High-level reward} $r^h_t$ at time $t$ is derived from three parts: the prediction error of the final ensemble model, as well as the number and complexity of the selected base models.

    \item \textit{Low-level state} $s^{l}_{t}$ at time $t$ includes all the information contained in $s^{h}_{t}$, along with the model selection information indicated by $a_{t}^{h}$.

    \item \textit{Low-level action} $a^{l}_{t}$ is the vector of weights assigned to the selected models at time $t$. 

    \item \textit{Low-level reward} $r^{l}_{t}$ at time $t$ consists of two components: the prediction error of the final ensemble model and the base reward that can be obtained through the weight search.

\end{itemize}


\subsection{Encoder for State Representation}

Before the RL design, we first introduce a state encoder module that focuses on extracting useful data representations as the state for the following agents.
The encoder architecture is illustrated in the left part of figure with input information from both building data and base models.
For building data, we adopt temporal convolutional networks (TCN) \cite{bai2018empirical} for representation learning.
The networks consist of a stack of three TCN blocks with a kernel size of 4 and dilation rates of 1, 2, and 4. 
This architecture is designed to ensure that the network possesses a sufficiently long receptive field and each layer captures temporal patterns at varying granularities.
Additionally, to better capture the interrelationships, we introduce a cross-attention module for effective feature fusion.
Specifically, we set:
$\mathbf{q} = \mathbf{H_q W_q}, \mathbf{k} = \mathbf{H_k W_k}, \mathbf{v} = \mathbf{H_v W_v}$.
Here, $\mathbf{H_q}=\tau_{x}(\mathbf{x_{t}})$ and $\mathbf{H_k}=\mathbf{H_v}=\tau_{u}(\mathbf{u_{t-1}}) \oplus \tau_{d}(\mathbf{d_{t}})$, where $\tau$ represent the TCN networks.
$\mathbf{W_q} \in \mathbb{R}^{d_{h} \times d_{h}}$, $\mathbf{W_k}$ and $\mathbf{W_v} \in \mathbb{R}^{(2 d_{h}) \times d_{h}}$ are learnable projection matrices for attention.
Then, the embedding is computed as: $e_{t}^{d} = \text{Softmax}(\mathbf{qk}^T/\sqrt{d_{h}})\mathbf{v}$, where $d_{h}$ is the hidden dimension.
Furthermore, we evaluate the prediction performance of base models on historical data as supplementary information for model selection.
Specifically, the mean squared error of each model at the last timestamp $t-1$ is recorded and embedded as: $e_{t}^{m} = \text{Embed}(\{\mathcal{L}_{t-1}(f_{i})\}_{i=1}^{N})$.
Thus, the state representation is defined as:
\begin{equation}
    s_{t} = e_{t}^{d} \oplus e_{t}^{m}.
\end{equation}


\subsection{High-level Agent}



In the initial phase of the HRL framework, the high-level agent $\pi^{h}$ performs model selection based on the state $s_{t}^{h} = s_{t}$.
The objective is to identify a subset of candidate models that are relevant and beneficial for the thermodynamic modeling of the target building.

\textbf{Action.}
We define the high-level action $a_{t}^{h}$ as a vector of binary values $[b_{1},...,b_{N}]$, where $b_{i}=1$ indicates that the model $f_{i}$ is selected. Conversely, if $b_{i}=0$, $f_{i}$ will not be considered in the model ensemble.

\textbf{Reward.}
The reward should indicate whether the model ensemble performs effectively, thereby guiding the agent's learning process.
Consequently, an intuitive reward is the forecasting error of the ensemble model $f_{t}^{ens}$ at time $t$, defined as $r_{t}^{loss} = -\mathcal{L}_{t}(f_{t}^{ens})$.
However, this reward design alone may lead the high-level agent to select an excessive amount of irrelevant models, which increases the training burden on the low-level agent.
Besides, the model ensemble $f_{t}^{ens}$ derived from an extensive combination of models is unlikely to accurately reflect the real building environment due to the complex factors and features involved \cite{picard2017impact}.
To encourage the high-level agent to select a reasonably smaller amount of models with simpler structures, we introduce two internal rewards $r_{t}^{mod} = -\sum_{i=1}^{N} b_{i}$ and $r_{t}^{var} = -||\sum_{i=1}^{N}b_{i}f_{i}||$, which penalize the number of base models and variables selected.
Here $||\cdot||$ measures the amount of variables in a model.
Then, the high-level reward can be represented as:
\begin{equation}
    r_{t}^{h} = r_{t}^{loss} + \alpha \cdot r_{t}^{mod} + \beta \cdot r_{t}^{var},
\end{equation}
where $\alpha$ and $\beta$ adjust the influence of internal rewards.

\subsection{Low-level Agent}


The low-level agent $\pi^{l}$ performs ensemble weight search for selected models based on the state representation and the high-level action, i.e., $s_{t}^{l} = (s_{t}, a_{t}^{h})$.
The objective is to refine the contributions of selected models such that the model ensemble best approximates the real-world environment of the target building.

\textbf{Action.}
We define the low-level action $a_{t}^{l}$ as a vector of weights $[w_{1},...,w_{N}]$, where $w_{i} = 0$ if $b_{i} = 0$ and $w_{i} \in [0,1]$ if $b_{i} = 1$.
Since the indoor temperature typically remains within a relatively consistent range across buildings, and given that the model library contains sufficiently diverse base models, we enforce the constraint $\sum_{i=1}^{N} w_{i} = 1$ to narrow the action space for efficient weight search without significantly compromising prediction accuracy.

\textbf{Reward.}
The low-level agent is designed to search for the optimal weights with respect to base models selected by the high-level agent.
Considering different candidate models given by $\pi^{h}$, the reward that can be achieved by $\pi^{l}$ depends on the base performance of these models.
Therefore, we introduce the prediction performance of a base model ensemble, i.e., $r_{t}^{base} = -\mathcal{L}_{t}(\sum_{i=1}^{N}b_{i}f_{i} / \sum_{i=1}^{N}b_{i})$, to represent the base reward that can be achieved by the low-level agent without too much effort for weight search.
Specifically, this model ensemble is constructed by assigning all the selected models the same weights.
Thus, the low-level reward is defined as:
\begin{equation}
    r_{t}^{l} = r_{t}^{loss} - r_{t}^{base}.
\end{equation}

\subsection{Training}
We implement Multi-layer Perceptrons for the high-level and low-level agents $\pi^{h}$ and $\pi^{l}$, and adopt the REINFORCE algorithm \cite{williams1992simple} to leverage feedback from the environment for model selection and weight search guidance.
The agent parameters $\phi^{h}$ and $\phi^{l}$ are updated using the following gradients:
\begin{equation}
\begin{aligned}
\label{eq: gradient}
    & \bigtriangledown_{\phi}J = \mathbb{E}_{s_{t}, a_{t}}[\bigtriangledown_{\phi} \text{log}\pi(s_{t}, a_{t}|\phi) r_{t}]. \\
\end{aligned}
\end{equation}
To stabilize the training process, 
we update the agent by a linear combination of parameters of its old version and current version such that the parameters can change slowly and stably.
The update process is as follows:
\begin{equation}
\label{eq: update}
    \phi = \lambda \phi + (1-\lambda) \phi_{old}.
\end{equation}
Here, $\lambda$ is a coefficient that balances the contribution of two versions of the network and $\phi_{old}$ denotes the old agent.
Eq. \ref{eq: gradient} and \ref{eq: update} are applicable to both agents and we omit the superscript $^{h}$ and $^{l}$ for simplicity.

Additionally, considering the training difficulty of HRL, we propose a two-stage training process.
At the first stage, the low-level agent is pre-trained with randomly generated binary vectors in place of $a_{t}^{h}$, serving as the model selection results.
At the second stage, the high-level and low-level agent are jointly trained.
The objective of this design is to first optimize the low-level agent such that it can produce near-optimal combination weights no matter what model selection results it receives.
With a pre-trained low-level agent, we can mitigate the occurrence of situations where suitable base models are selected but unsatisfactory weights are assigned, which could adversely affect the learning of both agents.

\section{Offline Experiments}





We evaluate the performance of \textit{ReeM} in an offline setting, where the real traces of state $\mathbf{x}$, control input $\mathbf{u}$, and disturbance $\mathbf{d}$ are collected from real building environments. 
We address the following research questions:
\begin{itemize}

\item Q1: How does \textit{ReeM} perform in the building thermodynamics modeling task?

\item Q2: How do different components of \textit{ReeM} contribute to the decision-making in our context? 

\item Q3: How do hyperparameter settings affect performance?


\end{itemize}

\subsection{Evaluation Methodology}

\textbf{Building Environment Datasets.}
The data utilized for modeling were collected from 65 rooms in a research building at Osaka University, Japan. 
The floor area of rooms ranges from 27 to 179 $m^2$, and the presence of different equipment can impact the thermodynamics, e.g., a laboratory in the information science department contains more computers and workstations. 
In addition, the room location can also affect the thermodynamics, e.g., rooms situated on the easternmost or southernmost sides of the building are more susceptible to direct solar irradiation.
We let the sensor reading of indoor temperature as $\mathbf{x}$, instantaneous power of HVAC equipment\footnote{HVAC equipoment in this paper: Hitachi RAS-AP280SSR} as $\mathbf{u}$, and outdoor temperature as $\mathbf{d}$.
The data collection period spans four months, from November 2, 2023, to February 26, 2024, for all rooms.
The sensor data frequency is 15 min, and our experiments are conducted at both 15 min and 1 hour to simulate different sensor frequency settings. 



\textbf{Base Models Preparation.}
As shown in figure, the ensemble approaches rely on a model library that comprises a set of base models.
For the preparation of these base models, we follow the traditional paradigm depicted in figure.
We employ two existing data-driven modeling methodologies, resulting in two customized models for each room:
(i) Symbolic Regression (SR): This method aims to identify a mathematical expression to represent the relationships between variables and is commonly used in this context, such as for one-day-ahead prediction~\cite{leprince2021fifty} and 15-minute-ahead prediction~\cite{zhao2023data}.
(ii) Multiple Linear Regression (MLR): A more classical modeling approach in thermodynamics, which we re-implement ~\cite{moretti2021multiple} where the variables involved are the installed sensors and day type.
Figure presents the results of directly applying these base models. Notably, no single model consistently performs well across different rooms over the two-day evaluation.


\textbf{Evaluation Metrics.}
We evaluate the prediction performance using widely accepted metrics for this task, including MAE, MSE.

\textbf{Baselines.}
We compare \textit{ReeM} with the following baselines:
(1) Heuristics method. We first compare \textit{ReeM} with a classical heuristic method commonly used in dynamic model ensemble problems in data streaming scenarios~\cite{cerqueira2017dynamic}. We implement a simple scheme that leverages the best $N$ models at previous time step $t-1$ as the ensemble, using average weighting. In our experiments, $N=3$.
(2) RLMC~\cite{fu2022reinforcement}. This is an RL-based scheme utilizing DDPG, an actor-critic algorithm for dynamic model combination in time-series forecasting.
(3) DMS~\cite{feng2019reinforcement}. This is another Q-learning-based RL scheme for dynamic model selection.
(4) BOA~\cite{deng2024towards}. We also compare \textit{ReeM} with an ensemble method in the building energy domain. Unlike the aforementioned RL-based ensemble methods, BOA generates a static ensemble formulated as a Bayesian optimization problem and uses  Gaussian processes (GP) to search the optimal combination.

\textbf{Setup.}
We divided the total of 65 rooms into two subsets: 80\% (52 rooms) for base model construction and RL training, and 20\% (13 rooms) for evaluation.
The 52 rooms are utilized to develop 52 × 2 = 104 base models with the two traditional methods mentioned earlier.
Then, their data are used for RL training for our method and baselines\footnote{It is important to note that data preparation method does not lead to overfitting, as the customized models are not perfectly accurate for their corresponding rooms.}.
For the 13 test rooms, a 1:1 split was used, i.e., the last two months' data are used as the test set, while the previous two months' data are employed to customize the base models and to generate the ensemble for the BOA approach.
For HRL agents training, Adam optimizers are employed with learning rates of 0.001 and 0.0005 for the first and second training stage.
The hidden dimension $d_{h}$ of state representation is set as 64.
For the high-level agent, the internal reward weights $\alpha$ and $\beta$ are set as 0.005 and 0.0015 respectively.
The coefficient $\lambda$ for stable update is set as 0.001.

\begin{table}[]
    \centering
    \scalebox{0.75}{
    \begin{tabular}{l|cc|c|cc|c}
        \toprule
        \multirow{2}{*}{\textbf{Model}} & \multicolumn{3}{c|}{Sampling: 15 min} & \multicolumn{3}{c}{Sampling: 1 hour} \\ \cline{2-7}
        
        & MAE & MSE & $\textit{Imp.}$  & MAE & MSE & $\textit{Imp.}$  \\ 
        \midrule
        \textit{customized}-SR  & 0.957 & 5.302 & - & 0.962 & 6.060 & -  \\
        \textit{customized}-MLR  & 0.983 & 6.950 & - & 1.348 & 10.996 & -  \\
        \midrule
        Heuristic & 1.072 & 9.141 & -10.5\%  & 1.422 & 18.437 & -23.1\%  \\  
        BOA & 1.064 & 13.666 & -9.7\%  & 1.192 & 15.429 & -3.2\%  \\  
        RLMC & 1.047 & 13.420 & -7.9\%  & 1.462 & 21.479 & -26.6\%  \\
        DMS & 1.131 & 17.246 & -16.6\% & 1.890 & 49.865 & -63.6\% \\  
        \midrule
        *\textbf{\textit{ReeM}} & 0.538 & 2.101 & +44.5\%  & 0.572 & 2.880 & +50.5\%  \\
        \bottomrule
    \end{tabular}
    }
    \caption{Overall performance.}
    \label{tab: overall performance}
\end{table}


\subsection{Overall Performance (Q1)}

Table~\ref{tab: overall performance} presents the overall performance of all prediction methods across the 13 tested rooms, along with the performance improvements of the five ensemble methods compared to the two traditional modeling approaches.
The results indicate that \textit{ReeM} outperforms all the baselines across multiple metrics.
Notably, we observe that the four baseline ensemble methods perform significantly worse than a customized model tailored for the test room. This highlights the challenge of building an ensemble that achieves generalization for this task. Furthermore, the three dynamic ensemble methods (Heuristic, RLMC, DMS) do not consistently outperform the static ensemble method, BOA.
In comparison to RLMC and DMS, which also utilize RL for model weighting, our \textit{ReeM} incorporates a hierarchical architecture along with a tailored rewards design, enabling it to manage a larger number of base models (whereas the original RLMC considered only nine base models).
We also look into the detailed performance of the five ensemble methods for the individual rooms as shown in figure. 
In average, \textit{ReeM} achieves a better performance of 51.40\%, 50.94\%, 50.99\%, and 53.27\% to BOA, Heuristic, RLMC, DMS, respectively. 
We observe that \textit{ReeM} performs slightly worse than other two baselines in two rooms (1 floor and 6 floor machine room). This is 
attributed to the fact that these two rooms exhibit less variation in indoor temperature, remaining relatively stable between
22.2$^{\circ}\mathrm{C}$ to 24.6$^{\circ}\mathrm{C}$.

\begin{table}[]
\centering
 \scalebox{0.8}{
\begin{tabular}{l|c|c}
\toprule
\textbf{Task} & \textbf{MAE} & \textbf{$\text{Imp.}$} \\
\midrule
\textit{ReeM} & \textbf{0.538} & \textbf{+44.54\%} \\
\midrule
\quad -- Hierarchy agents (single-tier RL) & 1.065 & -9.79\% \\
\quad -- Rewards design ($r^l=r^h=-\mathcal{L}(f_{t}^{ens})$) & 0.986 & -1.65\% \\
\quad -- state-control representation & 0.851 & +12.27\% \\
\bottomrule
\end{tabular}
}
\caption{Ablation tests.}
\label{tab: ablation}

\end{table}

\subsection{The Ablation Study (Q2)}

We conducted ablation tests to to verify the effectiveness of different components in \textit{ReeM}.
Table~\ref{tab: ablation} presents the results of three ablation settings: 
i) hierarchy RL agents design is essential. Here we replace the proposed two-level agents to single-tier RL (similar to RLMC) as ablation settings.
ii) $r=-\mathcal{L}(f_{t}^{ens})$ indicates that two agents share same reward, i.e., the prediction accuracy of $f_{t}^{ens}$. It implies the effectiveness of the separated rewards design.
iii) the data representation (TCN for temporal relation extraction and attention for feature field interaction) are indispensable in our approach. In this setting, we replace TCN and attention mechanisms with fully-connected layers.



\subsection{The Study of Key Hyperparameters (Q3)}
%


We explore the effects of the two key hyperparameters on the model’s performance.
\textit{i) Effects of the Base Model Number.}
We adjusted the number of base models from 10, 20, and 30 up to 104 (the total number of models). The results presented in figure indicate that the proposed \textit{ReeM} achieves optimal performance at 30 models, with performance declining from 40 models onward. We attribute this phenomenon to the relative insufficiency of training data for \textit{ReeM} to effectively explore such a large action space. For comparison, we conducted the same study in RLMC, where performance decreased from 10 models.
\textit{ii) Effects of the Type of base model.}
We vary the three types of the base models in $\{ \text{SR}, \text{MLR},  \text{SR+ MLR}\}$ and compared their performance. The results are presented in figure. It was observed that the type of base model has minimal impact; specifically, the proposed \textit{ReeM} does not heavily rely on any particular modeling method for the base model.

\section{On-site Case Study}

To evaluate the proposed \textit{ReeM} framework's performance in real-world HVAC control optimization, we conducted an on-site heating experiment in February 2025 in two of the above studied rooms in Osaka University.
The room temperature and the HVAC energy/power consumption were experimentally verified 4 times from February 7 to 10, 2025. 
The thermal comfort requirement was maintaining the room temperature at 20$^{\circ}$C $\pm$ 2$^{\circ}$C from 10:00 a.m. to 20:00 p.m. 
To meet this requirement, in the experiment using SR-based model predictive control~\cite{zhao2023data}, the operation time, operation mode, and the temperature setpoint were determined by the MPC.
In the experiment with our proposed \textit{ReeM}, the operation time and temperature setpoint were both determined by the framework.
Compassion experiments with SR and the \textit{ReeM} based HVAC management framework were conducted on February 8 and 9, 2025, from 0:00 to 24:00. 
The previously described frameworks were deployed on the edge server\footnote{Edge server: WAGO 752-9401} and operated the HVAC system.
For the objective of the MPC control optimization, we follow ~\cite{zhao2023data}: $J_{all} = J_{comfort} + J_{consume} + P_e $, where $J_{comfort}$ is thermal comfort term, $J_{consume}$ is energy consumption term, and $P_{e}$ is the penalty term, including penalties for temperature and power exceeding limits~\footnote{More details about this case of HVAC control optimization design can be found in ~\cite{zhao2023data} for interested readers.}. 
Note that, other factors can be involved depends on the scenario, e.g., the electrical price change.

As shown in figure, both SR and \textit{ReeM} effectively control and maintain the room temperature within the required range during the target period. Additionally, Fig.~\ref{fig:online-exp}(c) and (d) indicate that the average HVAC power consumption of \textit{ReeM} is slightly lower than that of the customized SR under an automation process.
These experimental results confirm that the \textit{ReeM} approach can provide an accurate and efficient thermodynamic model for HVAC operation optimization.
Moreover, \textit{ReeM} enables real-time online updates of the thermodynamic model without requiring re-training, a breakthrough for dynamic systems where adaptability is critical, yet SR model updates typically involve computationally expensive retraining, which struggles to keep up with rapid environmental changes.

\section{Discussion and Limitation}

In this paper, we study the problem of developing the building thermodynamics model from a model ensemble perspective, leveraging existing models to create a new model tailored to the target building environment.
We propose an HRL-based model ensemble approach, \textit{ReeM}, which formulates the ensemble task as a two-level decision-making process: the high-level task determines whether a base model should be selected, while the low-level task focuses on assigning weights to the selected models.
We evaluated \textit{ReeM} against two types of customized models and four model ensemble baselines across multiple rooms. Additionally, we conducted an on-site case study to demonstrate its benefits in real-world HVAC system control.

Despite the superior performance of the proposed approach, 
\textit{ReeM} performs ensemble learning at a superficial level, resulting in two major limitations: 
First, the features of the base models (e.g., the involved sensors) must be unified, which limits the ensemble to rely solely on commonly shared features. When base models are derived from crowdsourcing, they may exhibit diverse structures and cannot be directly integrated into the ensemble due to feature misalignment.
Second, the ensemble process may introduce physical inconsistencies, such as combining a model operating in cooling mode with another in heating mode, as \textit{ReeM} prioritizes prediction accuracy without considering the underlying physical principles.
Future work should study automatically exploring the internal logic and mechanisms of the base models to develop a more reliable and generalizable ensemble.



\newpage

\newpage

\bibliographystyle{named}
\bibliography{ijcai25}

\end{document}